\documentclass{article}
\usepackage{spconf,amsmath,epsfig}
\usepackage{times}
\usepackage{epsfig}
\usepackage{graphicx}
\usepackage{amsmath}
\usepackage{amssymb}
\usepackage{pdfpages}
\usepackage{graphicx}
\usepackage{subcaption} 
\usepackage{array,multirow}
\usepackage{xspace}
\usepackage{siunitx}
\usepackage[export]{adjustbox}

\usepackage{tabularx,booktabs}
\newcolumntype{Y}{>{\centering\arraybackslash}X}
\usepackage{caption} 
\usepackage[export]{adjustbox}
\usepackage{subcaption} 
\usepackage[font=small,skip=5pt]{caption}
\usepackage{url}            

\begin{document}\sloppy

\setlength{\abovedisplayskip}{2.8pt}
\setlength{\belowdisplayskip}{2.8pt}

\newlength{\bibitemsep}\setlength{\bibitemsep}{.2\baselineskip plus .05\baselineskip minus .05\baselineskip}
\newlength{\bibparskip}\setlength{\bibparskip}{0pt}
\let\oldthebibliography\thebibliography
\renewcommand\thebibliography[1]{%
  \oldthebibliography{#1}%
  \setlength{\parskip}{\bibitemsep}%
  \setlength{\itemsep}{\bibparskip}%
}


\makeatletter
\DeclareRobustCommand\onedot{\futurelet\@let@token\@onedot}
\def\@onedot{\ifx\@let@token.\else.\null\fi\xspace}

\def\eg{\emph{e.g}\onedot} \def\Eg{\emph{E.g}\onedot}
\def\ie{\emph{i.e}\onedot} \def\Ie{\emph{I.e}\onedot}
\def\cf{\emph{c.f}\onedot} \def\Cf{\emph{C.f}\onedot}
\def\etc{\emph{etc}\onedot} \def\vs{\emph{vs}\onedot}
\def\wrt{w.r.t\onedot} \def\dof{d.o.f\onedot}
\def\etal{\emph{et al}\onedot}
\makeatother
\newcommand{\erhao}{\fontsize{21pt}{\baselineskip}\selectfont}

\def\x{{\mathbf x}}
\def\L{{\cal L}}

{\onecolumn

\noindent \textbf{\erhao{Unsupervised Learning for Optical Flow Estimation Using Pyramid Convolution LSTM}}

\vspace{2cm}

\noindent {\LARGE{Shuosen Guan, Haoxin Li, Wei-Shi Zheng}}

\vspace{2cm}

\noindent Code is available at: \\
\ \ \ \ \ \ \ \ \ \ \ \ \url{https://github.com/Kwanss/PCLNet}

\vspace{1cm}

\noindent For reference of this work, please cite:

\vspace{1cm}
\noindent Shuosen Guan, Haoxin Li, Wei-Shi Zheng.
``Unsupervised Learning for Optical Flow Estimation Using Pyramid Convolution LSTM.''
In \emph{Proceedings of IEEE International Conference on Multimedia and Expo(ICME).} 2019.

\vspace{1cm}

\noindent Bib:\\
\noindent
@inproceedings\{PCLNet-icme2019,\\
\ \ \   title=\{Unsupervised Learning for Optical Flow Estimation Using Pyramid Convolution LSTM\},\\
\ \ \  author=\{Shuosen Guan, Haoxin Li, Wei-Shi Zheng\},\\
\ \ \  booktitle=\{Proceedings of IEEE International Conference on Multimedia and Expo(ICME)\},\\
\ \ \  year=\{2019\}\\
\}
}

\twocolumn

\title{Unsupervised Learning for Optical Flow Estimation Using\\
Pyramid Convolution LSTM}
%

\name{Shuosen Guan\textsuperscript{1,3}\hspace{10mm} Haoxin Li\textsuperscript{2,3}\hspace{10mm} Wei-Shi Zheng\textsuperscript{1,3,*}\thanks{* Corresponding author}}
\address{\textsuperscript{1}{School of Data and Computer Science, Sun Yat-sen University, China} \\
\textsuperscript{2}{School of Electronics and Information Technology, Sun Yat-sen University, China}\\
\textsuperscript{3}{Key Laboratory of Machine Intelligence and Advanced Computing, Ministry of Education, China} \\
\small \texttt{guanshs@mail2.sysu.edu.cn},\quad \texttt{lihaoxin05@gmail.com},\quad \texttt{wszheng@ieee.org}}

\maketitle
\begin{abstract}

Most of current Convolution Neural Network (CNN)
based methods for optical flow estimation focus on learning
optical flow on synthetic datasets with groundtruth, which is
not practical. In this paper, we propose an unsupervised optical
flow estimation framework named PCLNet. It uses pyramid
Convolution LSTM (ConvLSTM) with the constraint of
adjacent frame reconstruction, which allows flexibly estimating
multi-frame optical flows from any video clip. Besides,
by decoupling motion feature learning and optical flow representation,
our method avoids complex short-cut connections used in existing frameworks while improving accuracy of optical flow estimation. Moreover, different from those
methods using specialized CNN architectures for capturing motion, our framework directly learns optical
flow from the features of generic CNNs and thus can be easily
embedded in any CNN based frameworks for other tasks.
Extensive experiments have verified that our method not only estimates
optical flow effectively and accurately, but also obtains
comparable performance on action recognition.


\end{abstract} 
\begin{keywords}
Optical flow estimation, motion representation, unsupervised learning
\end{keywords}
\section{Introduction}
\label{sec:intro}
As a key problem in video analysis, optical flow estimation is widely used in lots of fields such as visual SLAM, autonomous driving, action recognition, \etc. Traditional optical flow estimation methods (\eg TVL-1~\cite{10.1007/978-3-540-74936-3_22}) treat the estimation problem as a energy function minimization problem, making strong assumptions on the pixel-level information such as brightness and spatial smoothness~\cite{Horn:1981:DOF:3015381.3015388}. These methods are time-consuming and are hard to use for large scale video data. 

With the remarkable success of CNN in image understanding, the optical flow estimation problem  has been promoted by recently proposed deep learning based methods such as FlowNet~\cite{Dosovitskiy2015}, FlowNet2.0~\cite{Ilg2017}, PWC-Net~\cite{Sun2017a},~\etc. These methods are end-to-end trainable and meanwhile achieve better accuracy and lower time consumption than the traditional methods. However, most of these CNN based methods need groundtruth optical flow for supervision during learning, which is almost unavailable in lots of real-world applications, such as action recognition on website videos, pedestrian tracking on surveillance videos,~\etc.

In addition, these methods generally adopt a ``coarse-to-fine" manner to predict optical flow by utilizing  auxiliary short-cut connections among high-level semantic layers. This brings the following shortcomings: First, it increases the complexity on the short-cut connections handling; Second, with this coupled architecture binding motion feature learning and optical flow representation, these frameworks make it hard to extend to other tasks. Moreover, these methods tend to customize  architectures for optical flow estimation and thus generally need training from scratch on some synthetic datasets, which leads to low compatibility with currently mature CNN backbones (\eg VGG, ResNet,~\etc).

We address the above issues in this work from three aspects: First, for the sake of unsupervised learning on real-world videos, we formulate the optical flow estimation as a reconstruction problem using adjacent frames and employ the reconstruction loss~\cite{Yu} to guide our network training. Second, our framework decouples motion capturing and optical flow representation by first applying spatial pyramid pooling to aggregate multi-scale motion information into low dimension motion features and then decoding optical flow directly from these features. As a result, it avoids complex connections while improving estimation accuracy. Last, our method constructs optical flow learning on generic CNNs and thus can be flexibly embedded in other CNN based architectures for other tasks, meanwhile, significantly reduce the time for network training with ImageNet pre-training. Different from other unsupervised learning methods~\cite{Yu, Zhu2017}, our method performs optical flow learning using ConvLSTM rather than merely CNN. Therefore, it is able to  simultaneously handle multi-frames on video sequences.

In summary, the contributions of our work are of three folds: First,  we propose an unsupervised  approach to simultaneously learn multi-frame optical flow on real-world videos without groundtruth.\ Second, our method decouples motion learning and optical flow representation.\ In this way, it avoids time consumption on complex short-cut connections among high-level semantic layers while improving estimation accuracy. Last, our model is compatible with generic CNNs and thus can be  embedded in general CNN based frameworks.

\vspace{-2mm}
\section{Related Work}
\label{sec:relaw}

Traditional methods for flow estimation are mainly based on variational approach. The most representative one  is the method proposed by Horn and Schunck~\cite{Horn:1981:DOF:3015381.3015388}.  It estimates optical flow by minimizing an energy function with some photometry assumptions such as brightness consistency and spatial smoothness. However, these assumptions  could not stand in some realistic cases. Other methods make improvements by choosing approximative but looser constraints, \eg~\cite{Black:1996:REM:229144.229157}. Nevertheless, these methods prone to be  restricted due to the hand-crafted design of these assumptions.

FlowNet~\cite{Dosovitskiy2015} takes an significant step to first deal with this problem by using a simple U-Net~\cite{Ronneberger_2015} architecture to learn optical flow layer by layer in a refinement way. After that, many other CNN based optical flow estimation methods were proposed. Among them, FlowNet2.0 pushes this task forward by stacking several sub-models in FlowNet (FlowNetS and FlowNetC) into a large model and obtains great improvement on estimation. SpyNet~\cite{Ranjan2017} only calculates optical flow residual at each layer and thus obtains  smaller model size and faster inference speed than previous methods. PWC-Net~\cite{Sun2017a} constructs optical flow following three principles: pyramidal processing, warping and cost volume. By applying these principles at each CNN layers, PWC-Net achieved much better results than previous ones with less parameters. To capture motion information, these methods generally compute cost volume at different feature levels using a special CNN layer named ``Correlation", which takes a patch of the second feature as a convolution kernel to conduct convolution on the first feature. It can be viewed as a matching scheme to find regions with spatial similarities between two feature representations.

In general, these CNN based methods for optical flow estimation mainly focus on learning optical flow on the videos with groundtruth and use complex short-cut connections for refinement. By contrast, our method explores to learn optical flow on real-world videos without groundtruth data. Moreover, by applying ConvLSTMs to capture motion information on multi-scale and aggregate motion information before optical flow estimation, our framework is capable of learning  multi-frame optical flows without these complex  connections. Besides, different from those methods using ``correlation" layer to capture motion between two frames, we use ConvLSTM to implicitly learn the motion dynamics and decode it as optical flow. Thanks to the compatibility with generic CNNs,  our framework can be flexibly embedded in other CNN based networks.

\begin{figure*}[t]
\includegraphics[width=0.97\linewidth]{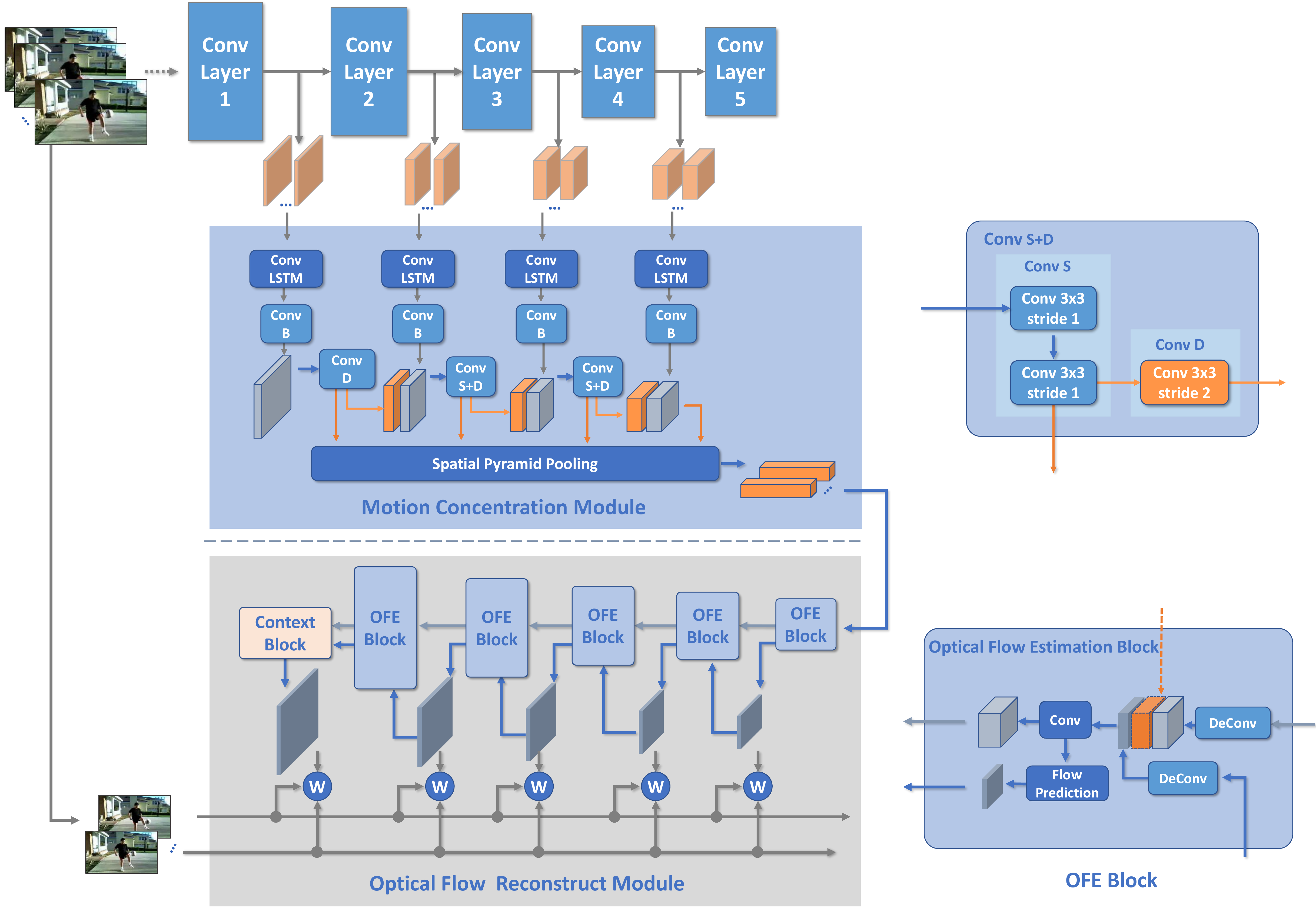}
  \caption{Our proposed framework, mainly including three modules. Feature Extraction Head (top): generic CNNs;  Motion Concentration Module (middle): learns motion features by pyramidal ConvLSTMs and then encodes the hidden states into low dimension motion features. Optical Flow Reconstruction Module (bottom): reconstructs optical flows from the learned motion features. Note that here ``W" denotes ``inverse warp" operation. The ``OFE Block" means ``Optical Flow Estimation Block" (bottom right), and the  orange dotted bordered cube in it  only exists in PCLNetC (Section~\ref{sec:abl_experiments}) and is removed in PCLNet.
}
\label{fig:framework}
\vspace{-3mm}
\end{figure*}

\vspace{-1mm}
\section{Approach} 
Our framework mainly consists of three modules: the generic CNN that used for appearance feature extraction, the motion concentration module that learns multi-scale motion representation and the optical flow reconstruction module that estimates optical flows from the motion features (see Figure~\ref{fig:framework}). We use the generic ResNet18~\cite{He_2016} as our feature extraction network and take the output of  its 4 blocks for the following processing. In the remaining of this section, we present the details about the other two modules:  motion concentration module and the optical flow reconstruction module. 

\vspace{-2mm}
\subsection{Motion Concentration Module}
\label{sec:mcm}
For completeness, we first give a brief introduction to ConvLSTM that we use for capturing motion dynamics on CNN feature pyramid. Like the original LSTM, ConvLSTM  uses four ``gates" to control data flowing and self-updating including input gate $i_t$, forget gate $f_t$, cell gate $c_t$ and output gate $o_t$.  It takes 2D feature maps as input and performs convolution operation  between them. The overall operation can be formulated as:
\begin{flalign}
i_t &= \sigma(W_{xi}\ast X_{t} + W_{hi}\ast H_{t-1} + W_{ci}\circ c_{t-1} + b_i)\notag\\
f_t &= \sigma(W_{xf}\ast X_{t} + W_{hf}\ast H_{t-1} + W_{cf}\circ c_{t-1} + b_f)\notag\\
c_{t} &= f_{t}\circ c_{t-1} + i_{t} \circ tanh(W_{xc}\ast X_{t} + W_{hc}\ast H_{t-1} + b_c)\notag\\
o_{t} &= \sigma (W_{xo} \ast X_{t} + W_{ho}\ast H_{t-1} + W_{co}\circ c_{t} + b_{0})\notag\\
H_{t} &= o_t \circ tanh(c_t)
\end{flalign}
where $\ast$ denotes convolution operation, $\circ$ denotes Hadamard product and $\sigma$ denotes $sigmoid$ operation. 

We apply a ConvLSTM layer to each output layer of the feature extraction CNN. Suppose that the length of each input clip is $l$. Then the output hidden state of each ConvLSTM layer has the shape of $l\times{c_i}\times{h_i}\times{w_i}$, where  $c_i,h_i,w_i$ is the output channel, height and width of the corresponding $i^{th}$ layer respectively. We take the latter $l-1$ hidden states as the learned motion expression among these $l$ frames. After that, we use several convolution layers to refine the learned hidden features and pass them to the next layer. Finally, to capture the motion information on different scales, we apply Spatial Pyramid Pooling~\cite{he2014spatial} to the output of all layers to generate fixed-length low-dimension motion features.


\subsection{Optical Flow Reconstruction Module}
\label{sec:ofrm}
To constrain the model to learn motion dynamics, we reconstruct optical flows among adjacent frames of each input clip from the learned motion features. Since in most cases, the groundtruth optical flow is not available for realistic tasks, inspired by~\cite{Yu,Zhu2017}, we formulate the optical flow estimation problem as an image reconstruction problem.

Let $I=\{I_1, I_2\}$ be two adjacent frames as input, and then the transformation from $I_1$ to $I_2$ can be formulated as $I_2=W(I_1, F)$, where $F$ is the optical flow from $I_1$ to $I_2$, and $W$ means warping operation that maps each point of $I_1$ to $I_2$ according to the flow $F$. Conversely, given the estimated flow $\hat{F}$, we can reconstruct $I_1$ as $\hat{I_1} = W^{-1}(I_2, \hat{F})$ using the inverse warping operation $W^{-1}$~\cite{Jaderberg2015}. To obtain the estimated optical flow $\hat{F}$, we minimize the similarity error between $\hat{I_1}$ and $I_1$, which can be formulated as
\begin{equation}
\hat{F} = min\  L_{sim}(\hat{I_1}, I_1),
\end{equation}
where $L_{sim}$ is similarity loss function. Here, we use three kinds  of similarity loss function as follows. \\
{\noindent\bf Pixelwise Similarity Loss.} It directly computes the pixelwise error between two frames using the generalized Charbonnier penalty function to alleviate the effects of outliers:
\begin{equation}
L_{diff} =\frac{1}{N}\sum\limits_{i,j}^{N}( (\hat{I_1}(i,j) - I_1(i,j))^{2} +\epsilon)^{\alpha},
\label{eq:loss_diff}
\end{equation}
where $\alpha$ is the Charbonnier penalty factor.\\
 {\noindent\bf Modified Peak Signal-to-Noise Ratio (PSNR) Loss.} PSNR  measures the peak error between two images. It is a common pixel difference-based measure of picture degradation. Here, we modified the original PSNR to be greater than $0$ :
\begin{equation}
L_{prsn} =  -10 \cdot \log_{10} \left(\frac{1}{1 + MSE(\hat{I_1}, I_1)}\right), 
\end{equation}
where $MSE$ denotes the mean square error measure. \\
{\noindent\bf Mean Structural Similarity (MSSIM) Loss.}  Structural Similarity (SSIM)~\cite{Wang2004} considers image degradation as perceived change of three aspects: luminance, contrast and structure information. It is based on the assumption that closer regions in space have stronger inter-dependencies and provide structure information of image appearance. Overall similarity error is computed by first splitting the reference and target images into small windows and then computing the mean SSIM (MSSIM) error over all windows: \\
\begin{equation}
L_{ssim} = 1- \frac{1}{M}\sum\limits_{m}^{M}SSIM(\hat{{I_1}}_m,\ {I_1}_m),
\label{eq:loss_sim}
\end{equation} 
where $SSIM(.)$ is the standard Structural Similarity function and $M$ denotes the number of windows we split.

Finally, we take the weighted sum of these three similarity loss functions as the final reconstruction loss:
\begin{equation}
L_{rec} = {\beta}_1 \cdot L_{diff} + {\beta}_2 \cdot L_{prsn}  + {\beta}_3 \cdot L_{ssim}
\label{eq:loss_all}
\end{equation} 
Here, ${\beta}_1 $, ${\beta}_2$ and ${\beta}_3$ are the balanced weights of these three similarity loss, respectively.

To reconstruct optical flow in higher resolution, we upscale the predicted flow by stacking five Optical Flow Estimation (OFE) blocks (bottom right in Figure~\ref{fig:framework}) from bottom to top. Here, the ``Flow Prediction" in OFE blocks is a simple convolution layer that shrinks the feature channel to $2$ without changing the height and width. We compute the reconstruction loss at each layer and add up to the final unsupervised reconstruction losses.\ Besides, to capture spatial context for finer flow representation, we apply  a ``Context Block" consisting of 7 dilated convolution layers on the top as described in~\cite{Sun2017a}. Note that we refine the optical flows merely from the low dimension motion features without utilizing  auxiliary short-cut connections from high-level features.

\section{Experiments}
\subsection{Datasets}
{\noindent\bf Datasets without groundtruth.} We investigate the performance of optical flow estimation on two real-world action recogniton datasets: UCF101~\cite{Soomro2012} and HMDB51~\cite{Kuehne_2011}. UCF101 consists of 101 action categories and 13,320 videos. HMDB51 contains 6766 videos clips from 51 action classes. 

{\noindent\bf Datasets with groundtruth.} We perform experiments on two synthetic optical flow datasets: FlyingChairs~\cite{Dosovitskiy2015} and MPI-Sintel~\cite{Butler:ECCV:2012}. FlyingChairs consists of 22,872 frame pairs and is split into 22,232 training examples and 640 testing examples. MPI-Sintel consists of 1628 frames with higher resolution and diverse motion patterns. It is split into training set and testing set. Both of them are divided into three ``passes": Albedo, Clean and Final. To supervise our training, we further split the training set into training split and validation split as \cite{Ranjan2017}.  

\begin{figure}[t]
\begin{center}
\includegraphics[width=0.92\linewidth]{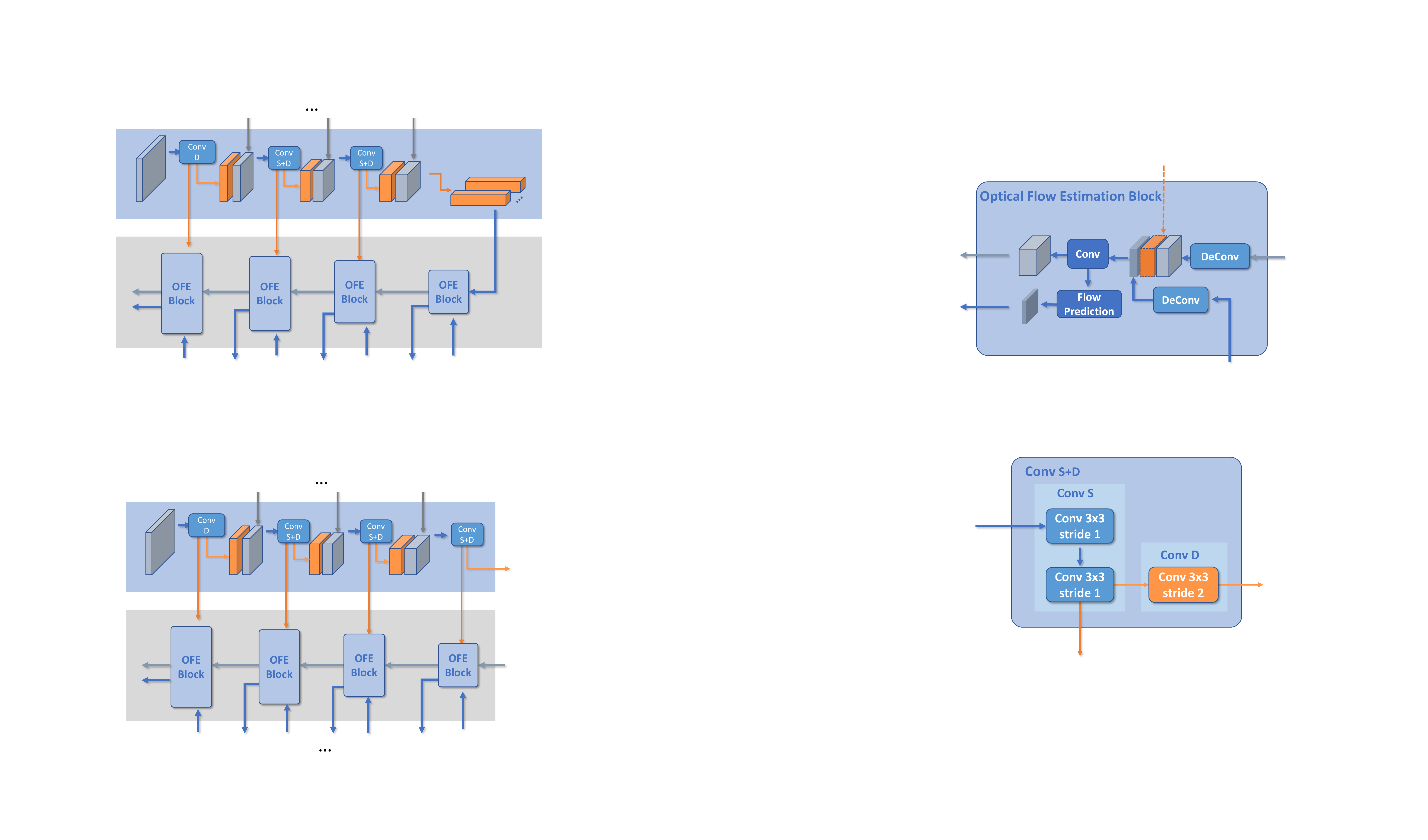}
\end{center}
   \caption{Illustration of the ``couple connection" in PCLNetC (denoted as orange arrows from top to bottom). Note that except the coupled connections, others are exactly the same as in PCLNet (see Figure~\ref{fig:framework}). }
\label{fig:block_couple_link}
\end{figure}

\subsection{Training Details}
We first pretrain our network on FlyingChairs and MPI-Sintel dataset. Specifically, We initialize the feature extraction network (ResNet18 in our experiments) with the weights pretrained from ImageNet. In this stage, we use the average end-point-error (EPE) as the training loss. Here, we use single GPU and set batch size as 32 for training. We start training with an initial learning rate of $1\mathrm{e}{-4}$ and then decay the learning by 0.5 at step 60K, 80K, 100K and 120K.  For data augmentation, we apply random crop, random scale, horizontal flipping and vertical flipping in the optical flow estimation task. Note that, we perform augmentation jointly for each input clip to keep the optical flows coherent.

Next, we train our model on HMDB51 and UCF101 dataset without groundtruth optical flow supervision. We also apply data augmentation as in the pretraining stage. We split each video into multiple clips. The size of each input clip is set to $6\times 224\times 224$. Note that in this stage, we only use  reconstruction loss for training and set the $\alpha$ as $0.4$ (Eq.~\ref{eq:loss_diff}), $M=7$ (Eq.~\ref{eq:loss_sim}), and $\beta_1 = 1.0, \beta_2 = 0.2, \beta_3 = 0.5 $ (Eq.~\ref{eq:loss_all}) based on validation. Here, we set batch size as 48 and start training with a learning rate as $1\mathrm{e}{-3}$, and then decay the learning by 0.1 when the training loss plateaus. 
All of the above experiments were conducted on NVIDIA Titan V GPUs.
\vspace{-5mm}

\subsection{Ablation Study}
\label{sec:abl_experiments}
To investigate the impact on our ``decouple" design and  compare with those methods using short-cut connections, we design another U-net like PCLNet, which is illustrated in Figure~\ref{fig:block_couple_link}. We denote this network as ``PCLNetC" in the following section. In this model, we add a couple connection between each layer in motion concentration module and the corresponding layer in the optical flow reconstruction module. The remaining part of this model is the same as in PCLNet.
\smallskip
\begin{table}
\begin{center}
\caption{Comparison of the EPE (End-point-error) and runtime between PCLNetC and PCLNet on MPI-Sintel dataset (training set).}
\label{tb:ablation}
\resizebox{\columnwidth}{!}{%
\begin{tabular}{l|cccc|c}
\hline
Models & Clean & Final & Clean(ft) & Final (ft) & Time (s)  \\ 
\hline 
PCLNetC & 5.71 	& 7.58 & 4.15	& 5.37 & 0.056	\\
PCLNet  & 5.52  & 6.87 & 3.69   & 4.39 & 0.052	\\
\hline
\end{tabular}
}
\end{center}
\vspace{-6mm} 
\end{table}

We evaluate these two kinds of models on the training set of MPI-Sintel dataset.\ The result is shown in Table~\ref{tb:ablation}.\ Surprisingly, even using lighter framework, the PCLNet without ``couple connections" achieves better result than the coupled one. We own this improvement to the employment of  spatial pyramid pooling: by integrating multi-scale information into low-dimension features before decoding as optical flows, it gains even better performance without using auxiliary information from high-level semantic layers. Moreover, this decoupled design reduces time consumption to a small degree.

\subsection{Comparison Results}
\label{sec:results}
\begin{figure*}[t]
     \centering
     \begin{adjustbox}{totalheight=5.58cm, width=0.85\textwidth, minipage=[s]{\linewidth}}
     
    \begin{subfigure}[t]{0.198\linewidth}
        \raisebox{-\height}{\includegraphics[width=\textwidth]{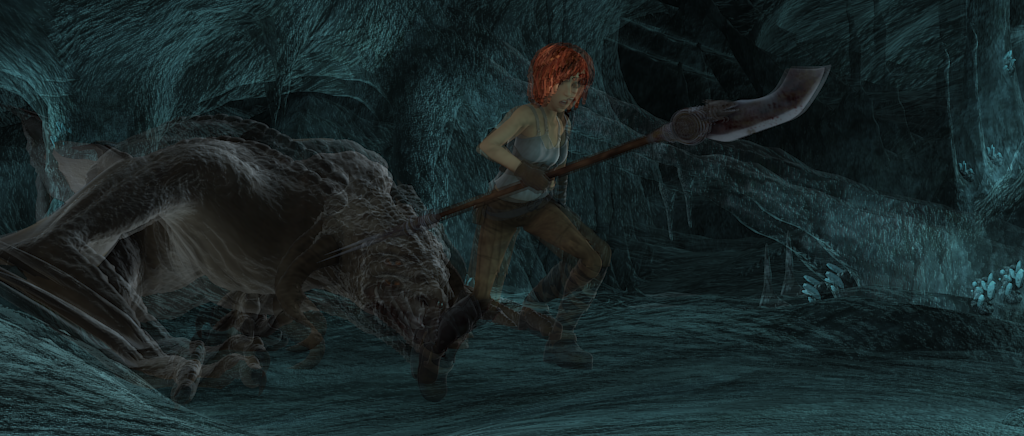}}
    \end{subfigure}\hspace{0.008\linewidth}\hfill%
    \begin{subfigure}[t]{0.198\linewidth}
        \raisebox{-\height}{\includegraphics[width=\textwidth]{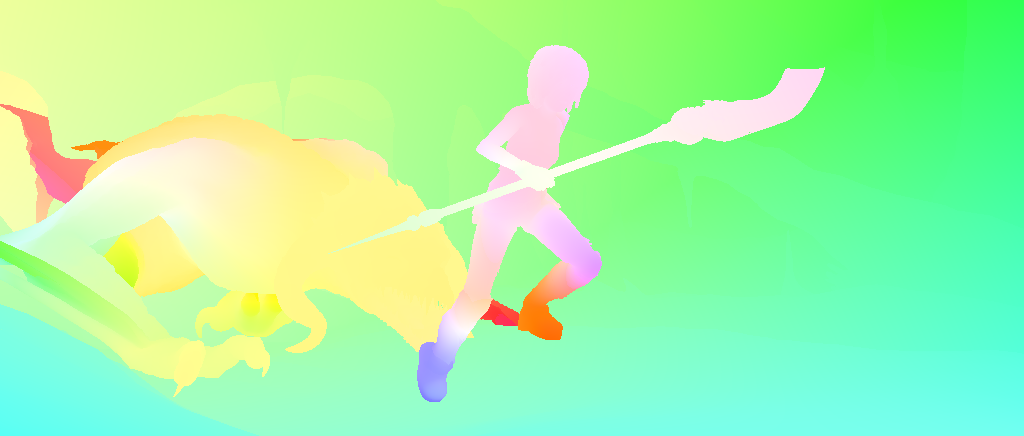}}
    \end{subfigure}\hfill%
    \begin{subfigure}[t]{0.198\linewidth}
        \raisebox{-\height}{\includegraphics[width=\textwidth]{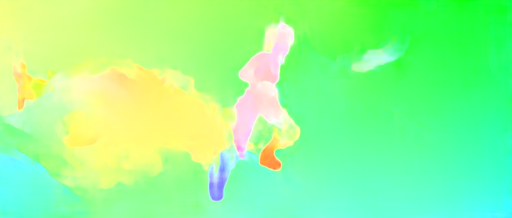}}
    \end{subfigure}\hfill%
    \begin{subfigure}[t]{0.198\linewidth}
        \raisebox{-\height}{\includegraphics[width=\textwidth]{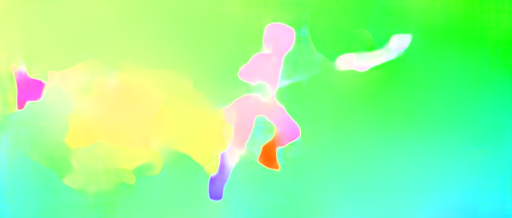}}
    \end{subfigure}\hfill%
        \begin{subfigure}[t]{0.198\linewidth}
        \raisebox{-\height}{\includegraphics[width=\textwidth]{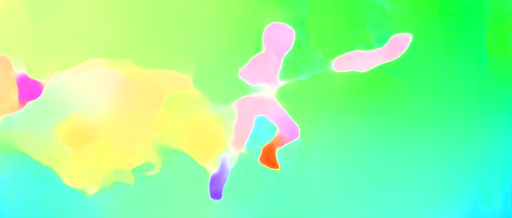}}
    \end{subfigure}
    
    \vfill%
    \begin{subfigure}[t]{0.198\linewidth}
        \raisebox{-\height}{\includegraphics[width=\textwidth]{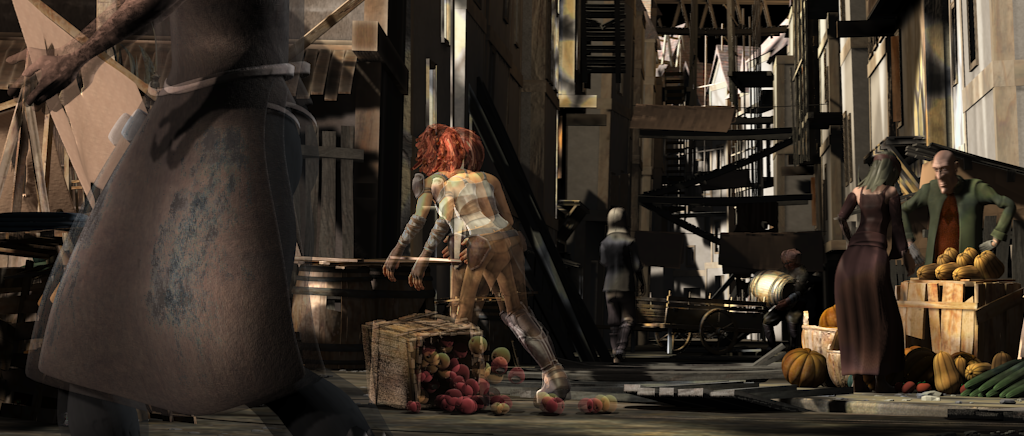}}
    \end{subfigure}\hspace{0.008\linewidth}\hfill%
    \begin{subfigure}[t]{0.198\linewidth}
        \raisebox{-\height}{\includegraphics[width=\textwidth]{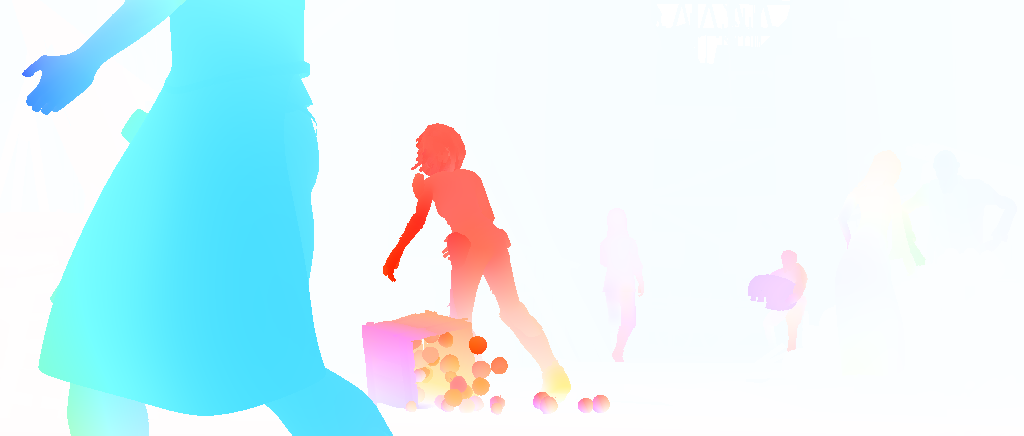}}
    \end{subfigure}\hfill%
    \begin{subfigure}[t]{0.198\linewidth}
        \raisebox{-\height}{\includegraphics[width=\textwidth]{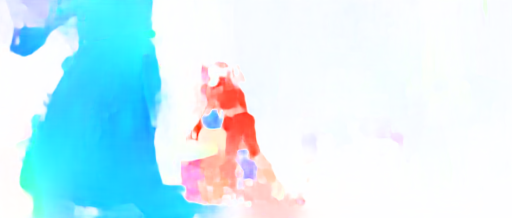}}
    \end{subfigure}\hfill%
    \begin{subfigure}[t]{0.198\linewidth}
        \raisebox{-\height}{\includegraphics[width=\textwidth]{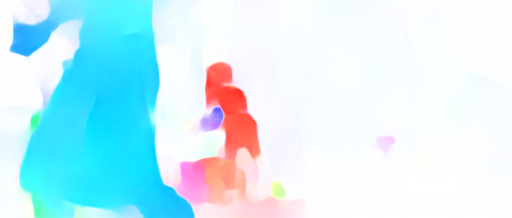}}
    \end{subfigure}\hfill%
        \begin{subfigure}[t]{0.198\linewidth}
        \raisebox{-\height}{\includegraphics[width=\textwidth]{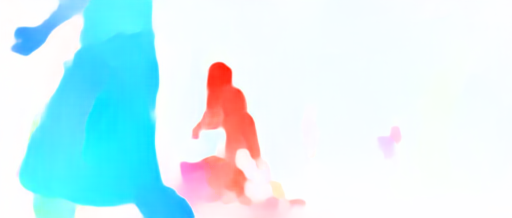}}
    \end{subfigure}
    
    \vfill%
    \begin{subfigure}[t]{0.198\linewidth}
        \raisebox{-\height}{\includegraphics[width=\textwidth]{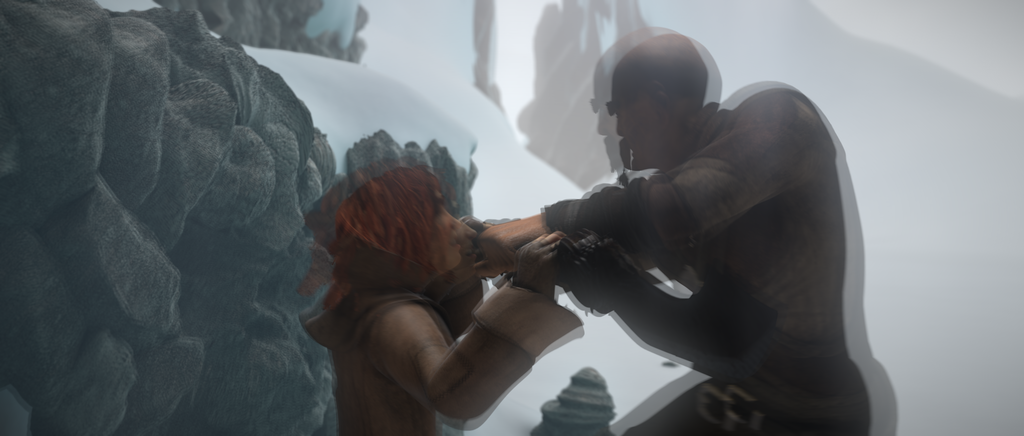}}
    \end{subfigure}\hspace{0.008\linewidth}\hfill%
    \begin{subfigure}[t]{0.198\linewidth}
        \raisebox{-\height}{\includegraphics[width=\textwidth]{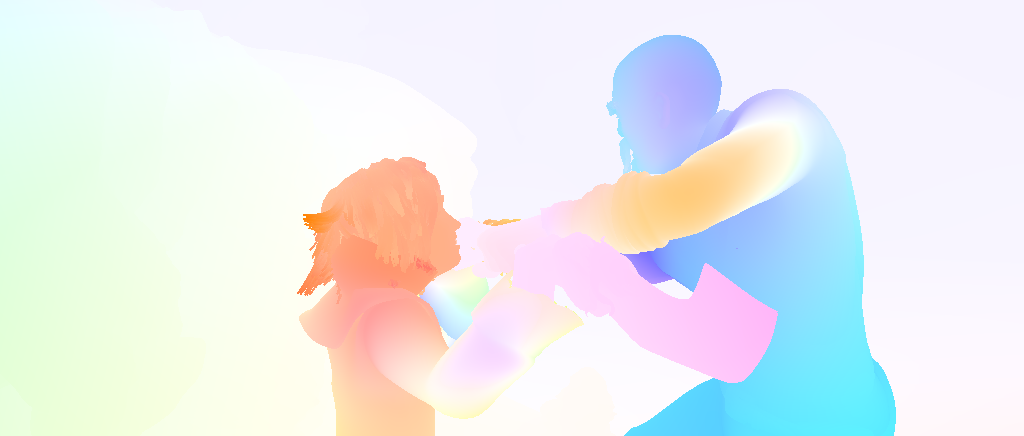}}
    \end{subfigure}\hfill%
    \begin{subfigure}[t]{0.198\linewidth}
        \raisebox{-\height}{\includegraphics[width=\textwidth]{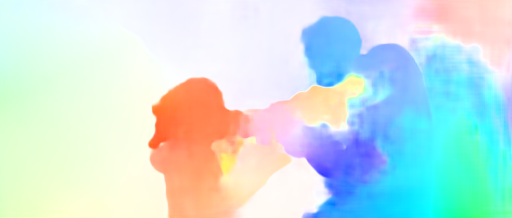}}
    \end{subfigure}\hfill%
    \begin{subfigure}[t]{0.198\linewidth}
        \raisebox{-\height}{\includegraphics[width=\textwidth]{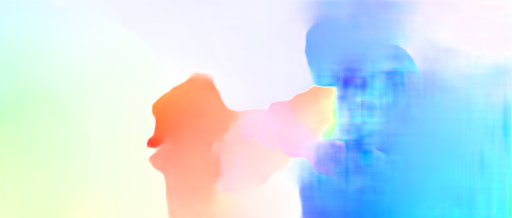}}
    \end{subfigure}\hfill%
        \begin{subfigure}[t]{0.198\linewidth}
        \raisebox{-\height}{\includegraphics[width=\textwidth]{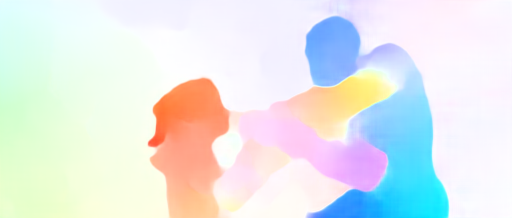}}
    \end{subfigure}

    \vfill%
    \begin{subfigure}[t]{0.198\linewidth}
        \raisebox{-\height}{\includegraphics[width=\textwidth]{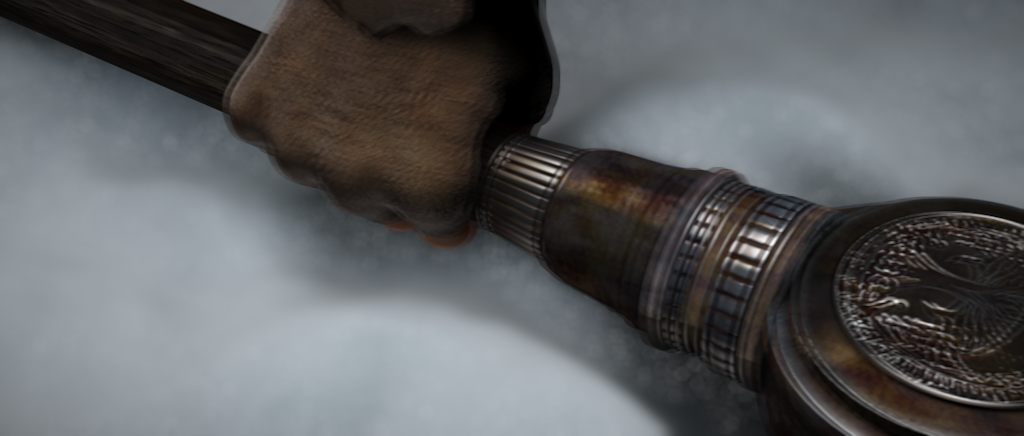}}
        \caption{frames overlay}
    \end{subfigure}\hspace{0.008\linewidth}\hfill%
    \begin{subfigure}[t]{0.198\linewidth}
        \raisebox{-\height}{\includegraphics[width=\textwidth]{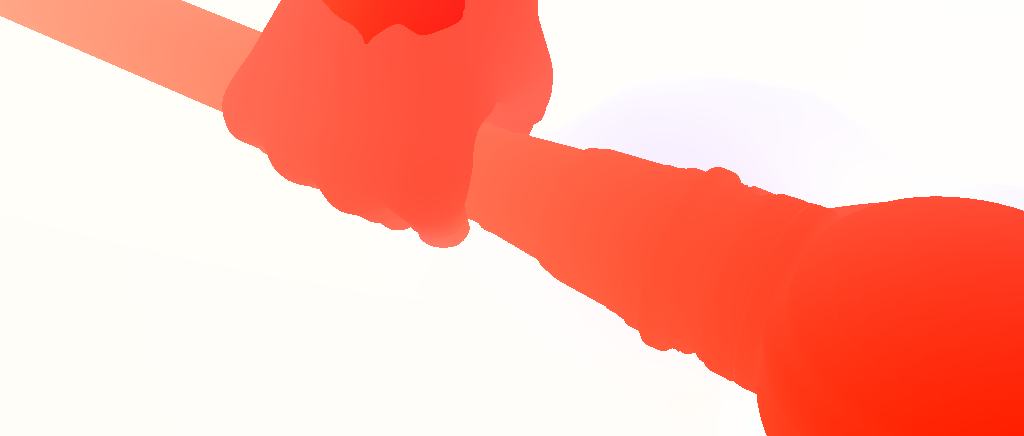}}
        \caption{ground truth}
    \end{subfigure}\hfill%
    \begin{subfigure}[t]{0.198\linewidth}
        \raisebox{-\height}{\includegraphics[width=\textwidth]{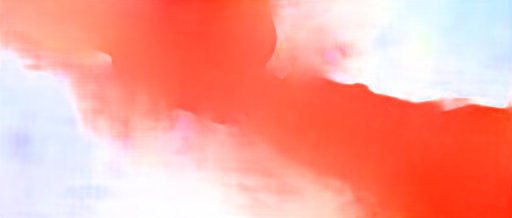}}
        \caption{PCLNetC}
    \end{subfigure}\hfill%
    \begin{subfigure}[t]{0.198\linewidth}
        \raisebox{-\height}{\includegraphics[width=\textwidth]{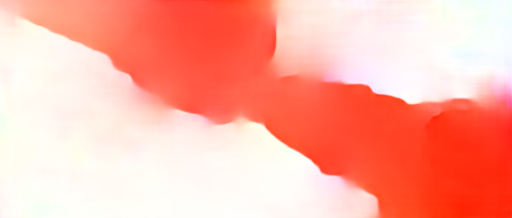}}
        \caption{PCLNet}
    \end{subfigure}\hfill%
        \begin{subfigure}[t]{0.198\linewidth}
        \raisebox{-\height}{\includegraphics[width=\textwidth]{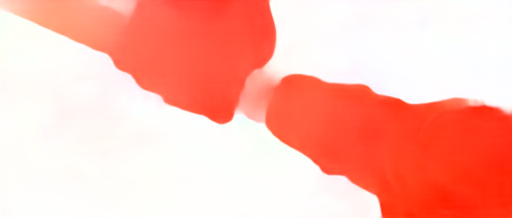}}
        \caption{PCLNet-ft}
    \end{subfigure}
     \end{adjustbox}
\caption{Visual results for optical flow estimation on MPI-Sintel dataset. Top two rows: results evaluated on Clean pass. Bottom two rows: results on Final pass. ``-ft" means that the model has been finetuned on the dataset.}
\label{fig:flow_visual_mpi}
\vspace{-4mm}
\end{figure*}

Obviously, our framework can be extended for entirely supervised learning by simply replacing the reconstruction loss (Eq.~\ref{eq:loss_all}) to EPE loss. For comparison, we first conduct an experiment to perform optical flow estimation on MPI-Sintel dataset with merely EPE loss. The results are shown in Table~\ref{tb:table_flow}. As shown, our method achieves better performance than traditional TVL-1 on both accuracy and efficiency. As for CNN based optical flow estimation methods, even though our model gains slightly worse accuracy, it achieves promising runtime performance comparing with other methods. It shows that although we estimate optical flow from the low-dimension motion features without high-level short-cut connections, our method still achieves a good trade-off between accuracy and efficiency. The visualized results of the estimated optical flows on both PCLNet and PCLNetC are shown in Figure~\ref{fig:flow_visual_mpi}. We can see that both of these two kinds of  models are able to estimate clear optical flows while the PCLNet is slightly better on capturing details. 


\begin{table}
\begin{center}
\caption{EPE and time consumption of different optical flow estimation methods evaluated on MPI-Sintel dataset. Here, ``ft" denotes that the method has been finetuned on the training split.}
\label{tb:table_flow}
\tabcolsep=0.17cm
\begin{tabular}{|c|cccc|c|}
\hline
\multirow{2}{*}{Method}   & \multicolumn{2}{c}{Training} &  \multicolumn{2}{c|}{Test} & Time \\
 & Clean & Final&Clean & Final & (s)\\
\hline \hline
\multicolumn{1}{|l|}{TV-L1~\cite{10.1007/978-3-540-74936-3_22}} & - & - & 9.471 & 10.462 & 0.068 \\
\hline
\multicolumn{1}{|l|}{FlowNetC~\cite{Dosovitskiy2015}} & 4.31 & 5.87 & 7.28 & 8.81 & 0.150 \\
\multicolumn{1}{|l|}{FlowNetC-ft} & 3.78 & 5.28 & 6.85 & 8.51 & 0.150 \\
\multicolumn{1}{|l|}{FlowNetS~\cite{Dosovitskiy2015} } & 4.50 & 5.45 & 7.42 & 8.43 & 0.080 \\
\multicolumn{1}{|l|}{FlowNetS-ft} & 3.66 & 4.44 & 6.96 & 7.76 & 0.080 \\
\multicolumn{1}{|l|}{FlowNet2.0~\cite{Ilg2017} }& 2.02 & 3.14 & 3.96 & 6.02 & 0.123 \\
\multicolumn{1}{|l|}{FlowNet2.0-ft}& 1.45 & 2.01 & 4.16 & 5.74 & 0.123 \\
\multicolumn{1}{|l|}{SpyNet~\cite{Ranjan2017} }  & 4.12 & 5.57 & 6.69 & 8.43 & 0.069 \\ 
\multicolumn{1}{|l|}{SpyNet-ft}  & 3.17 & 4.32 & 6.64 & 8.36 & 0.069 \\
\hline
\multicolumn{1}{|l|}{PCLNet}  & 5.52 & 6.87 & 8.79 & 9.63 & 0.052 \\
\multicolumn{1}{|l|}{PCLNet-ft}  & 3.69 & 4.39 & 7.92 & 8.58 & 0.052 \\
\hline
\end{tabular}
\end{center}
\vspace{-4mm} 
\end{table}

\begin{figure}[t]
     \centering
    \begin{subfigure}[t]{0.248\linewidth}
        \raisebox{-\height}{\includegraphics[width=\textwidth,height=0.68\textwidth]{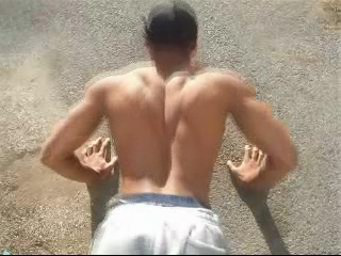}}
    \end{subfigure}\hfill%
    \begin{subfigure}[t]{0.248\linewidth}
        \raisebox{-\height}{\includegraphics[width=\textwidth,height=0.68\textwidth]{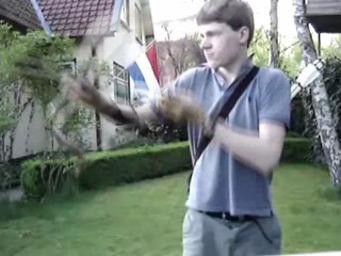}}
    \end{subfigure}\hfill%
    \begin{subfigure}[t]{0.248\linewidth}
        \raisebox{-\height}{\includegraphics[width=\textwidth,height=0.68\textwidth]{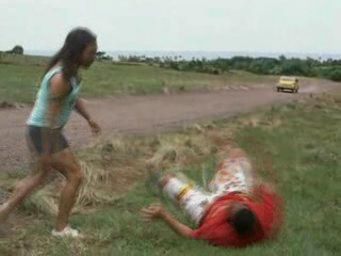}}
    \end{subfigure}\hfill%
    \begin{subfigure}[t]{0.248\linewidth}
        \raisebox{-\height}{\includegraphics[width=\textwidth,height=0.68\textwidth]{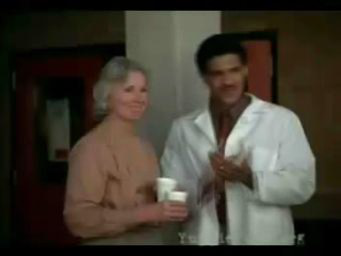}}
    \end{subfigure}

    \vfill%
    \begin{subfigure}[t]{0.248\linewidth}
        \raisebox{-\height}{\includegraphics[width=\textwidth]{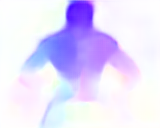}}
    \end{subfigure}\hfill%
    \begin{subfigure}[t]{0.248\linewidth}
        \raisebox{-\height}{\includegraphics[width=\textwidth]{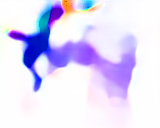}}
    \end{subfigure}\hfill%
    \begin{subfigure}[t]{0.248\linewidth}
        \raisebox{-\height}{\includegraphics[width=\textwidth]{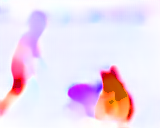}}
    \end{subfigure}\hfill%
    \begin{subfigure}[t]{0.248\linewidth}
        \raisebox{-\height}{\includegraphics[width=\textwidth]{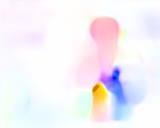}}
    \end{subfigure}

    \begin{subfigure}[t]{0.248\linewidth}
        \raisebox{-\height}{\includegraphics[width=\textwidth,height=0.68\textwidth]{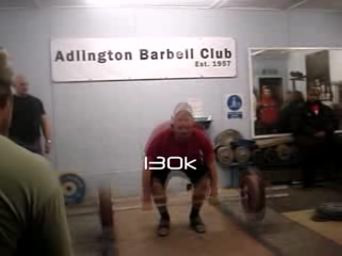}}
    \end{subfigure}\hfill%
    \begin{subfigure}[t]{0.248\linewidth}
        \raisebox{-\height}{\includegraphics[width=\textwidth,height=0.68\textwidth]{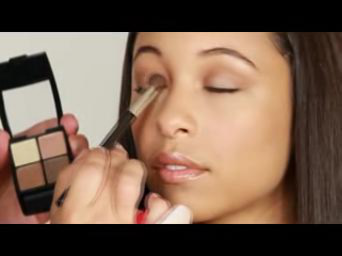}}
    \end{subfigure}\hfill%
    \begin{subfigure}[t]{0.248\linewidth}
        \raisebox{-\height}{\includegraphics[width=\textwidth,height=0.68\textwidth]{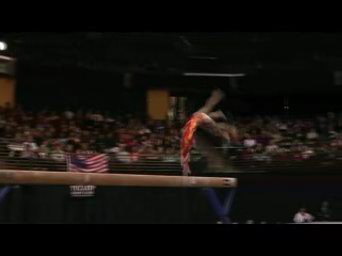}}
    \end{subfigure}\hfill%
    \begin{subfigure}[t]{0.248\linewidth}
    	\raisebox{-\height}{\includegraphics[width=\textwidth,height=0.68\textwidth]{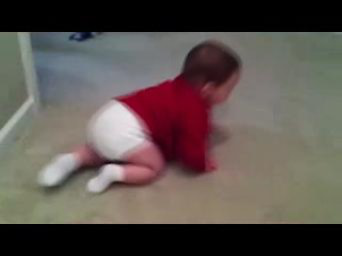}}
    \end{subfigure}

    \vfill%
        \begin{subfigure}[t]{0.248\linewidth}
        \raisebox{-\height}{\includegraphics[width=\textwidth]{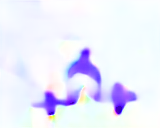}}
    \end{subfigure}\hfill%
    \begin{subfigure}[t]{0.248\linewidth}
        \raisebox{-\height}{\includegraphics[width=\textwidth]{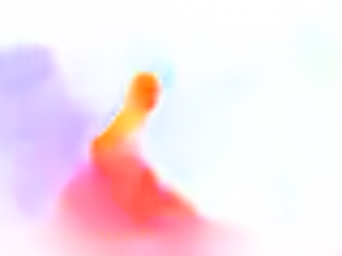}}
    \end{subfigure}\hfill%
    \begin{subfigure}[t]{0.248\linewidth}
        \raisebox{-\height}{\includegraphics[width=\textwidth]{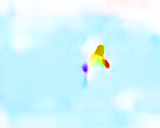}}
    \end{subfigure}\hfill%
    \begin{subfigure}[t]{0.248\linewidth}
    	\raisebox{-\height}{\includegraphics[width=\textwidth]{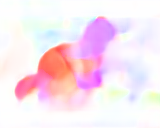}}
    \end{subfigure}
    
\caption{Visualization of the reconstructed optical flows from PCLNet on HMDB and UCF101 dataset. The top two rows are the frames overlays and corresponding estimated flow. The bottom two rows are examples from UCF101 respectively.}
\label{fig:flow_visual_cls}
\vspace{-2mm}
\end{figure}

As for unsupervised learning on videos without groundtruth, we evaluate our method on HMDB51 and UCF101 using PCLNet. The estimated optical flows are visualized in Figure~\ref{fig:flow_visual_cls}. As shown, our framework is able to estimate crisp optical flow merely using the reconstruction constraint. To quantitatively verify our method for optical flow estimation on real-world videos, we further perform an experiment on action recognition task on these datasets compared with other optical flow estimation methods including  TVL-1~\cite{10.1007/978-3-540-74936-3_22}, DIS-Fast~\cite{Kroeger_2016}, Deepflow~\cite{weinzaepfel2013deepflow}, FlowNet~\cite{Dosovitskiy2015} and  FlowNet2.0~\cite{Ilg2017}. Specifically, we first pre-compute the optical flows on these two datasets, and then we take them as input and use a ResNet50 network to  perform action recognition. The result is shown in Table~\ref{tb:table_cls_m}. Although our result is slightly worse than the results of TV-L1, we enjoy higher accuracy than other specialized motion descriptors.

\begin{table}
\begin{center}
\caption{Comparison of the classification accuracy of different motion descriptors on HMDB51(split 1) and UCF101 (split 1) datasets.}
\label{tb:table_cls_m}
\begin{tabularx}{0.8\linewidth}{|l|Y|Y|}
\hline
Method & HMDB51 & UCF101\\
\hline \hline
TV-L1~\cite{10.1007/978-3-540-74936-3_22} & 56.0 	& 83.9\\
DIS-Fast~\cite{Kroeger_2016}	& 40.4 & 71.2 \\
Deepflow~\cite{weinzaepfel2013deepflow}	& 50.4 & 82.1 \\
FlowNet~\cite{Dosovitskiy2015} 	& 38.6	& 55.3 \\
FlowNet2.0~\cite{Ilg2017}	& 52.3 & 80.1 \\
\hline
Ours	&53.5 & 82.8 \\
\hline
\end{tabularx}
\end{center}
\vspace{-6mm} 
\end{table}

\section{Conclusions}
In this paper, we present a novel end-to-end trainable framework for optical flow estimation. By utilizing reconstruction constraint as supervision, our framework is able to efficiently learn optical flow on real-world videos without groundtruth. In addition, we decouple motion feature learning and optical flow reconstruction by applying ConvLSTMs on multi-scale to capture motion dynamics and then decode the learned motion features into optical flows. In this way, our method simultaneously estimates multi-frame optical flow without complex short-cut connections. Furthermore, this framework is compatible with generic CNNs and thus can be easily embedded in CNN based Networks for other video analysis tasks. Experiments verified that our approach achieves  promising results for optical flow estimation on both synthetic and real-world video data.

\vspace{4mm} 
 {\noindent\bf\large{Acknowledgement}} This work was supported partially by the National Key Research and Development Program of China (2018YFB1004903), NSFC (61522115,U1811461), Guangdong Province Science and Technology Innovation Leading Talents (2016TX03X157), and the Royal Society Newton Advanced Fellowship (NA150459).

\vspace{-1mm}


\bibliographystyle{IEEEbib}
\bibliography{icme2019template}

\end{document}